\documentclass[journal ]{new-aiaa}
\usepackage[utf8]{inputenc}
\usepackage{textcomp}

\usepackage{graphicx}
\usepackage{amsmath}
\usepackage[version=4]{mhchem}
\usepackage{siunitx}
\usepackage{longtable,tabularx}
\usepackage{verbatim}
\usepackage{subcaption}
\usepackage{bm}
\usepackage{bbm}
\usepackage{threeparttable}
\usepackage{footnpag}

\title{Bayesian Deep Learning for Segmentation for Autonomous Safe Planetary Landing \footnote{This paper is a substantially revised version of Paper AAS 21-253 presented at the AAS/AIAA Space Flight Mechanics Meeting, Virtual, 2021.}}

\author{Kento Tomita \footnote{Ph.D. Student, Daniel Guggenheim School of Aerospace Engineering, AIAA Student Member.}}
\affil{Georgia Institute of Technology, Atlanta, GA, 30332}
\author{Katherine A. Skinner\footnote{Assistant Professor, Department of Robotics, AIAA Member.}}
\affil{University of Michigan, Ann Arbor, MI, 48109}
\author{Koki Ho\footnote{Associate Professor, Daniel Guggenheim School of Aerospace Engineering, AIAA Senior Member.}}
\affil{Georgia Institute of Technology, Atlanta, GA, 30332}

\begin{document}

\maketitle

\begin{abstract}
Hazard detection is critical for enabling autonomous landing on planetary surfaces. Current state-of-the-art methods leverage traditional computer vision approaches to automate the identification of safe terrain from input digital elevation models (DEMs). However, performance for these methods can degrade for input DEMs with increased sensor noise. In the last decade, deep learning techniques have been developed for various applications. Nevertheless, their applicability to safety-critical space missions has often been limited due to concerns regarding their outputs' reliability. In response to these limitations, this paper proposes an application of the Bayesian deep-learning segmentation method for hazard detection. The developed approach enables reliable, safe landing site detection by: (i) generating simultaneously a safety prediction map and its uncertainty map via Bayesian deep learning and semantic segmentation;  and (ii) using the uncertainty map to filter out the uncertain pixels in the prediction map so that the safe site identification is performed only based on the certain pixels (i.e., pixels for which the model is certain about its safety prediction). Experiments are presented with simulated data based on a Mars HiRISE digital terrain model by varying uncertainty threshold and noise levels to demonstrate the performance of the proposed approach.
\end{abstract}

\section{Introduction}
\lettrine{A}{utonomous} landing is identified as a medium-high priority for future NASA missions~\cite{Mustard2013, Martin2015}. Real-time hazard detection and avoidance (HDA) is a critical component for enabling autonomous landing. The current state-of-the-art method, developed for the Autonomous Landing Hazard Avoidance Technology (ALHAT) program, demonstrated potential for computer vision algorithms to automate online identification of safe or unsafe terrain from input Digital Elevation Models (DEMs)~\citep{Johnson2008, Huertas2010, Rutishauser2012, Trawny2015}, which can be generated from measurements from a flash Light Detection and Ranging (LiDAR) sensor during a mission. The Chang'e-3 lunar lander mission, the only space mission that has successfully demonstrated autonomous HDA, also explicitly evaluated hazards based on DEMs in the precise hazard avoidance phase \cite{Li2016}. However, DEM-based explicit evaluation of hazards is sensitive to severe sensor noise, navigation error, and missing data, which can lead to errors in safe landing site prediction.

Recent work has demonstrated the potential for deep neural networks (DNNs) to improve both accuracy and computational cost for onboard hazard detection~\citep{Tomita2020, Moghe2020}. These prior works take input DEMs to output a binary map labeling each pixel as safe or unsafe for landing. However, one caveat in machine learning methods is their output's reliability. Most such methods rely on gathering a representative dataset for training, and may not generalize well to test data that is outside of the training distribution. In these cases, a predicted safety map will still be returned without any indication of how reliable that prediction is based on the input DEM.

There has been recent interest in estimating network uncertainty in deep learning predictions to provide further insight on the reliability of the network prediction~\citep{Gustafsson2020, Guo2017, Kendall2017, Mukhoti2018, Scalia2020}. However, these works have not been evaluated for autonomous landing on planetary surfaces, and there is a lack of literature on how to translate output uncertainty estimates into the safety map generation.

This work focuses on the development and validation of Bayesian deep learning methods for hazard detection for landing on a planetary surface. The specific contributions of this work are (i) to perform semantic segmentation and uncertainty prediction for identifying safe landing sites from input DEMs, (ii) to develop an algorithm that takes in the safety predictions and uncertainty maps to produce an uncertainty-aware (and thus reliable) identification of safe landing sites, and (iii) to perform experiments for demonstration of developed methods on DEMs by varying sensor noise and the uncertainty threshold. 

\section{Background}
\subsection{Autonomous Hazard Detection}
The ALHAT program was launched by NASA in 2005 to develop advanced capabilities for spacecraft autonomous hazard detection and precise landing on lunar and planetary surfaces~\citep{Epp2007, Brady2007, Striepe2010}. The work developed during the ALHAT program is considered state-of-the-art in autonomous hazard detection for planetary landing. In particular, Ivanov et al.~\citep{Ivanov2013} developed a probabilistic method for autonomous hazard detection that considers the vehicle size, the vehicle-surface configuration during landing, and navigation errors. The algorithm searches over the potential footpad positions and lander orientations during landing to determine the slope of the lander. The slopes are computed based on where the footpads fall on the terrain, as well as the roughness of hazards underneath the lander based on the final lander pose, i.e. position and orientation. If there are unsafe configurations for the lander at a given point on the DEM, this point is labeled as unsafe for landing. This provides pixel-wise labeling of safe and unsafe landing points. The final output of the algorithm is a list of proposed safe landing sites, along with their coordinates and safety probability. This work was integrated on board the Morpheus vehicle and has been extensively tested in real flight scenarios. Flight tests demonstrated successful autonomous hazard detection and hazard relative navigation capabilities for the Morpheus vehicle over realistic lunar-like terrain~\citep{Rutishauser2012, Epp2014, Trawny2015}. With simplifying assumptions, the ALHAT algorithm developed by Ivanov et al. is capable of operating in real-time for efficient and effective hazard detection during landing. However, the performance of the algorithm can degrade with increased sensor noise, navigation error, and missing data. 
Building off of the success of the ALHAT project and Morpheus experiments, next-generation hazard detection and avoidance systems are currently under development through NASA's Safe \& Precise Landing - Integrated Capabilities Evolution (SPLICE) program~\citep{Restrepo2020}.

\subsection{Machine Learning for Safe Landing on Planetary Surfaces}
 Recent work has focused on developing machine learning-based approaches to aid in guidance, navigation, and control for spacecraft landing on planetary surfaces~\citep{Iiyama2020, Gaudet2020, Scorsoglio2020, Furfaro2018}. For scenarios where high-resolution maps are available prior to landing, these methods can learn guidance and control policies to target the selected landing site. Our work focuses on the perceptual task of hazard detection to determine safe and unsafe landing sites when the DEM is computed onboard to provide additional information to the guidance system. This capability is critical for applications where high-resolution maps are not available prior to landing, and for detection of small hazards such as rocks.

There are several prior works that leverage machine learning for detecting and labeling geological features such as rocks and craters on the surface of the moon ~\citep{Di2014, Emami2015, Cohen2016, Silburt2019, Wang2019, Downes2020}, Mars~\citep{Palafox2017}, or small bodies~\citep{pugliatti2021board}. Our proposed work instead focuses specifically on detecting safe sites for landing without explicitly detecting hazardous terrain features such as craters or rocks. Recently, \citet{Moghe2020} developed a DNN for detecting safe and unsafe areas in a DEM for automated hazard detection. Their work develops a novel loss function specifically designed to decrease the false positive (safe) rate, since false positive cases indicate that an area is predicted safe when it is actually unsafe, which can lead to catastrophic situations for autonomous landing operations. \citet{Tomita2020} also explored several DNN architectures for real-time hazard detection. In particular, their results demonstrated that a semantic segmentation DNN can outperform the current state-of-the-art ALHAT algorithm for the prediction of safe pixels when the input DEM has sensor noise. Our work builds on these prior works to include estimation of network uncertainty for the output binary safety map to provide a cue on the reliability of the network prediction. Our proposed approach leverages output uncertainty maps to enable uncertainty-aware safe site identification. Furthermore, we provide experiments to evaluate our approach with varying noise parameters for testing the network. 

\subsection{Uncertainty Estimation}
There are two main types of uncertainty that can be modeled: aleatoric uncertainty and epistemic uncertainty~\citep{Gal_Thesis2016}. 
\textit{Aleatoric uncertainty} is uncertainty due to sensor noise or motion noise~\citep{Gal_Thesis2016}, whereas \textit{Epistemic uncertainty}, which is also referred to as model uncertainty, includes uncertainty in the model structure and parameters~\citep{Gal_Thesis2016}. 
Estimating uncertainty through deep learning frameworks has been investigated in the literature~\citep{Kendall2015}. 
For learning-based frameworks, aleatoric uncertainty is the main source of uncertainty if the test distribution is well-aligned with the training distribution (which is a common assumption in classical machine learning problems). 
However, predictions with low aleatoric uncertainty may still be incorrect due to the epistemic uncertainty (e.g., uncertainty in the DNN parameters), especially if the test distribution contains samples that fall outside of the training distribution. 
Particularly, epistemic uncertainty is an important contributor to the overall uncertainty when training sets are small or limited in variability, or for safety-critical applications. 
Epistemic uncertainty can be estimated through approximate Bayesian inference techniques, such as ensembling~\citep{Lakshminarayanan2017} or Monte Carlo (MC) dropout~\citep{Kendall2015}. 
For the application of autonomous landing on planetary surfaces, it is critical to account for both aleatoric and epistemic uncertainty to ensure that the safety map estimate considers uncertainty due to sensor noise and uncertainty due to encountering test cases that fall outside of the training distribution. 
Thus, our work will leverage prior work to estimate aleatoric and epistemic uncertainty, in order to enable uncertainty-aware safety map generation for a planetary lander.

\section{Proposed Methodology}
\begin{figure}[t!]
\centering
\includegraphics[width=.98\textwidth]{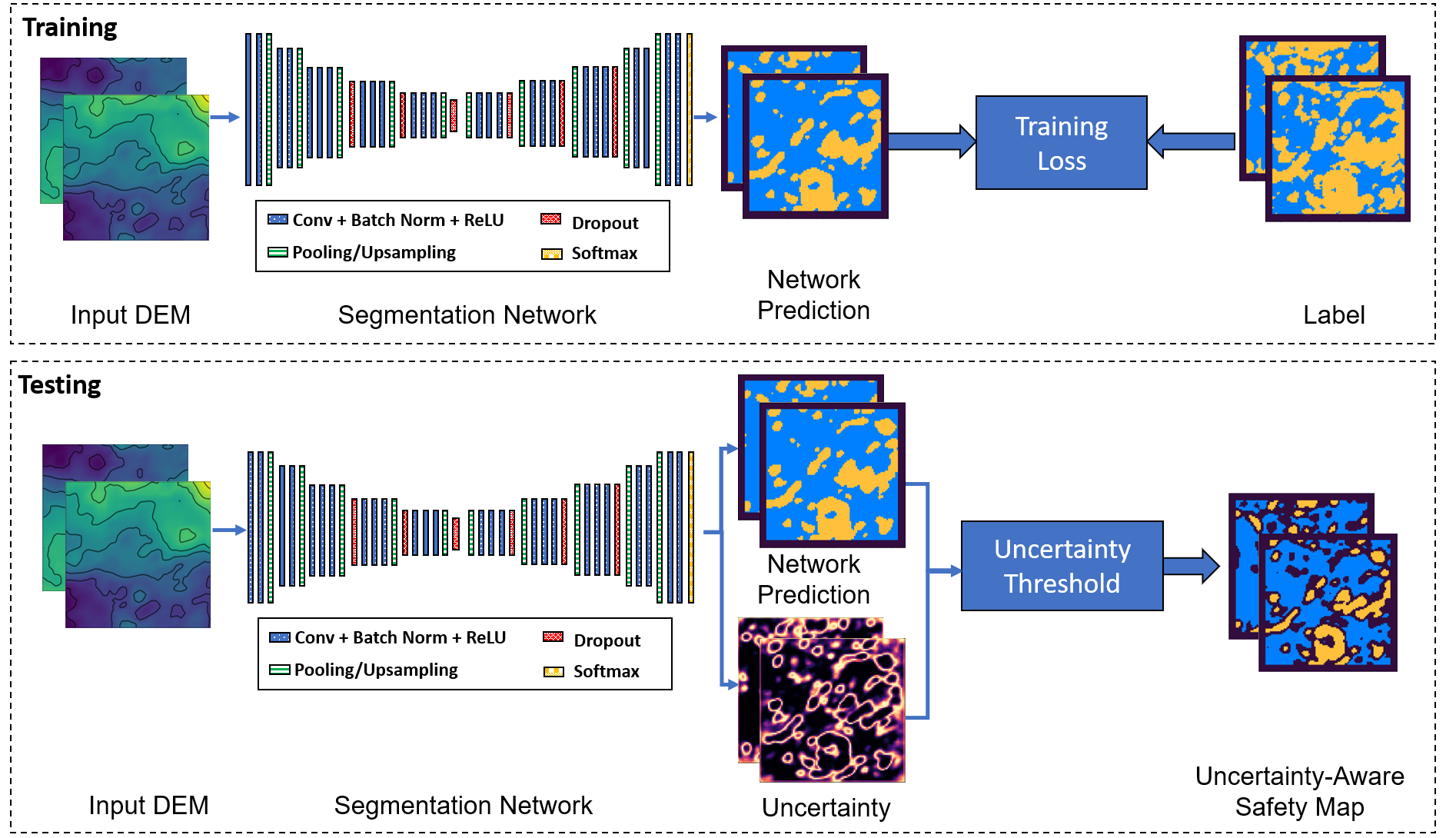}
\caption{Proposed approach for uncertainty-aware learning-based landing site detection.}
\label{fig:flowchart}
\end{figure}

Figure~\ref{fig:flowchart} shows an overview of our developed pipeline for learning-based safety map generation. The pipeline takes a noisy DEM obtained by noisy LiDAR scans as input to the semantic segmentation network. We assume that the raw output from LiDAR is converted to the DEM with associated information of the lander's position and attitude. The base network is the Bayesian semantic segmentation network, Bayesian SegNet~\citep{Kendall2015}. Bayesian SegNet leverages MC-dropout in which dropout is included during both training and testing. During inference, this allows for an output of both a pixel-wise safety prediction and an associated uncertainty that indicates how certain the network is in the prediction. The safety map and the associated uncertainty map are used to compute the uncertainty-aware safety map.

\subsection{Bayesian Deep Learning for Semantic Segmentation}
\begin{figure}[t!]
\centering
\includegraphics[width=.7\textwidth]{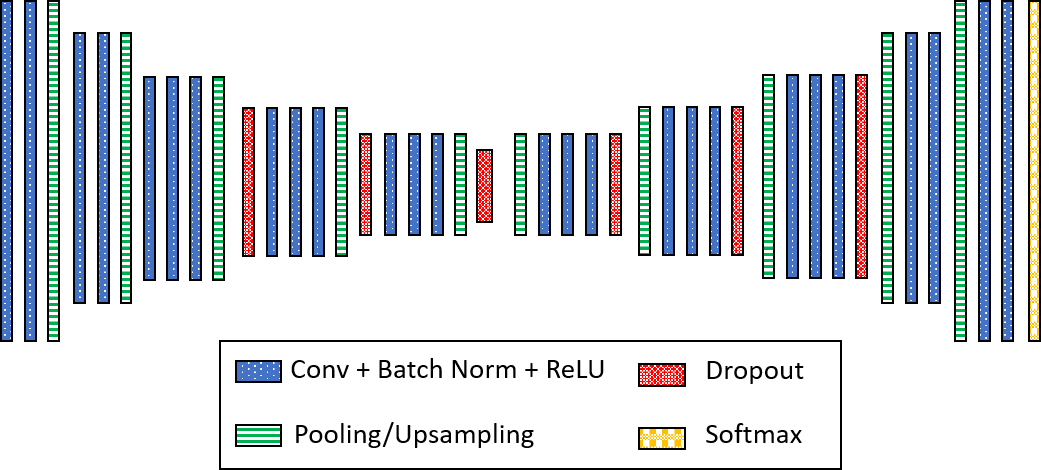}
\caption{A schematic of the network architecture for semantic segmentation, which is based on Bayesian SegNet~\cite{Kendall2015}.}
\label{fig:bsegnet}
\end{figure}

Figure~\ref{fig:bsegnet} shows the network architecture used for the semantic segmentation stage, which is based on Bayesian SegNet\cite{Kendall2015}. SegNet, the base architecture of Bayesian SegNet, is a state-of-the-art network for semantic segmentation~\cite{Badrinarayanan2017}. Our prior work demonstrated that SegNet performs well for hazard detection on planetary surfaces~\cite{Tomita2020}. The base structure of SegNet is a convolutional encoder-decoder, which successively downsamples the input until a bottleneck layer, at which point successive upsampling stages are used to output the desired resolution. Each convolutional layer includes convolution, batch normalization, and a rectified linear unit (ReLU) activation function. Convolutional blocks are followed by pooling/upsampling layers to achieve the encoder-decoder architecture. The network output is given as input to a softmax layer to produce the probability of each pixel to be labeled as safe or unsafe. The network is trained with cross-entropy loss.

For MC-dropout, network dropout layers are activated for both training and testing. During testing, each realization of the noisy DEM is input to the trained network $M$ times, where $M$ is the number of MC samples. The final prediction is given by the mean over the softmax output across the $M$ stochastic samples. Formally, let $\mathcal{D}_{train}$ be the training set, $\mathcal{C}=\text{\{Safe, Unsafe\}}$ be the set of pixel classes, $y$ be the output class associated with one DEM pixel $\bm{x}=(u, v)$, and $p(y=c|\bm{x}, \hat{w}_m)$ be the predicted probability that the DEM pixel $\bm{x}=(u, v)$ has the output label $c \in \mathcal{C}$ given the network weight realization $\hat{w}_m$ at the $m$th stochastic forward pass under random dropout. Then, the final prediction is

\begin{align}
\hat{p}(y=c|\mathbf{x},\mathcal{D}_{train}) = \frac{1}{M}\sum_{m=1}^{M} p(y=c|\mathbf{x},\hat{w}_m).
\end{align}

\noindent To output the final segmentation map, we take the label $c$ that corresponds to the highest $\hat{p}(y=c|\mathbf{x},\mathcal{D}_{train})$ for each pixel $\bm{x}$.

\subsection{Uncertainty Estimation}
The uncertainty map can be estimated during testing. There are several methods for computing the uncertainty map given the network output of the $M$ stochastic samples to capture aleatoric and/or epistemic uncertainty. These methods include computing the variance of the output samples, predictive entropy of the mean softmax probability, or mutual information between the predictive entropy and the posterior over model parameters~\citep{Mukhoti2018}. \citet{Mukhoti2018} notes that predictive entropy models both epistemic and aleatoric uncertainty, which are both important for safe planetary landing. Thus, for this work, we use predictive entropy to compute uncertainty maps. The predictive entropy, $\hat{\mathbbm{H}}[y|\mathbf{x},\mathcal{D}_{train}]$, is given by~\cite{Mukhoti2018}:

\begin{align}
    \hat{\mathbbm{H}}[y|\mathbf{x},\mathcal{D}_{train}] = - \sum_{c \in \mathcal{C}} \left[\left(\frac{1}{M}\sum_{m=1}^{M} p(y=c|\mathbf{x},\hat{w}_m)\right)\log\left(\frac{1}{M}\sum_{m=1}^{M} p(y=c|\mathbf{x},\hat{w}_m)\right)\right]
\end{align}

\begin{figure}[hbt!]
  \begin{subfigure}{0.17\textwidth}
    \includegraphics[trim=0 0 0 0, clip, height=32mm]{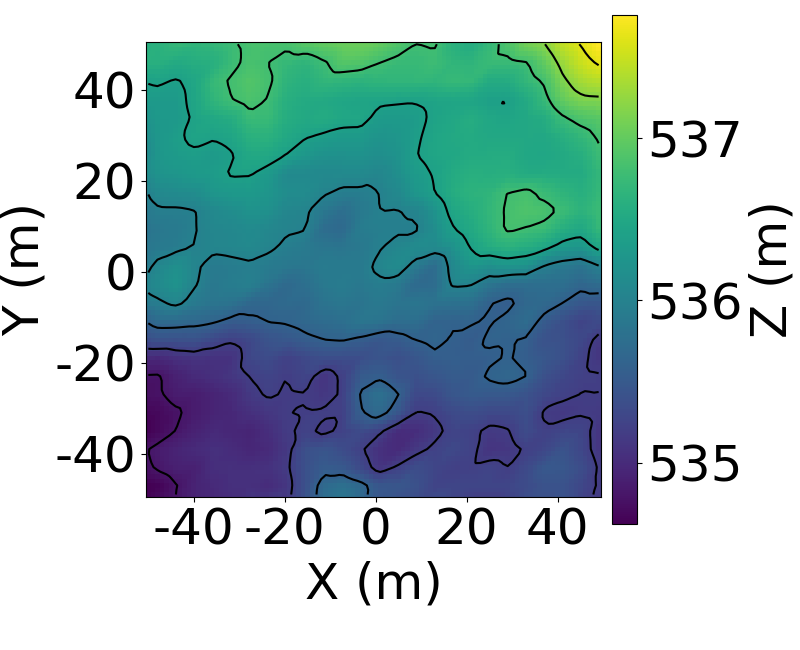}
    \caption{Input noisy DEM \newline} \label{fig:1a}
  \end{subfigure}%
  \hspace*{\fill}   
  \begin{subfigure}{0.15\textwidth}
    \includegraphics[trim=0 0 250 0, clip, height=30mm]{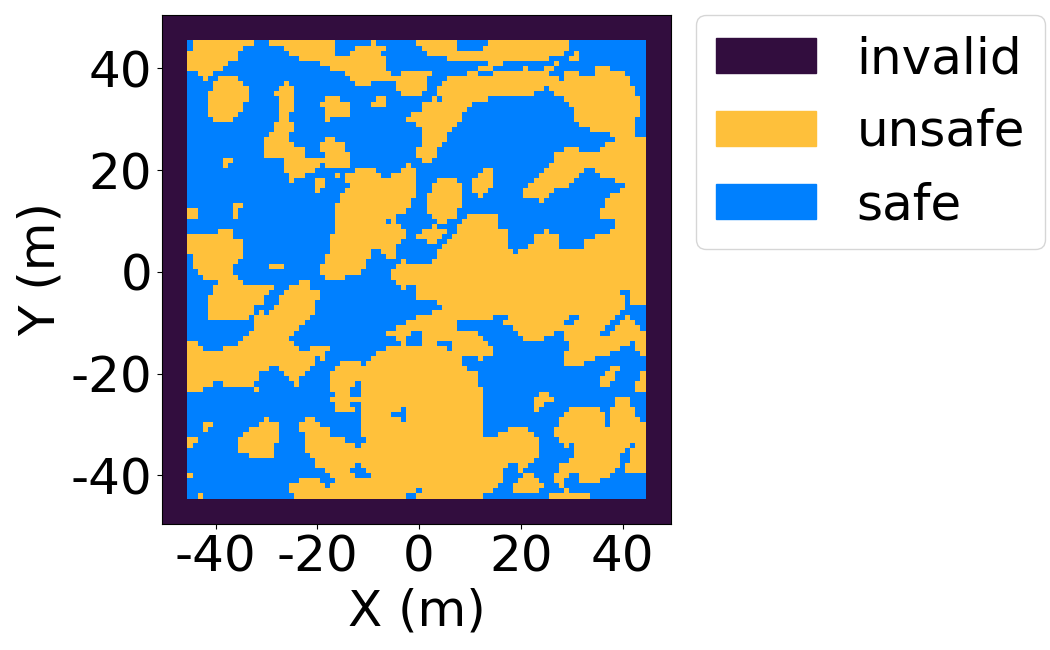}
    \caption{Ground Truth \newline} \label{fig:1b}
  \end{subfigure}%
  \hspace*{\fill}   
  \begin{subfigure}{0.20\textwidth}
    \includegraphics[trim=0 0 0 0, clip, height=30mm]{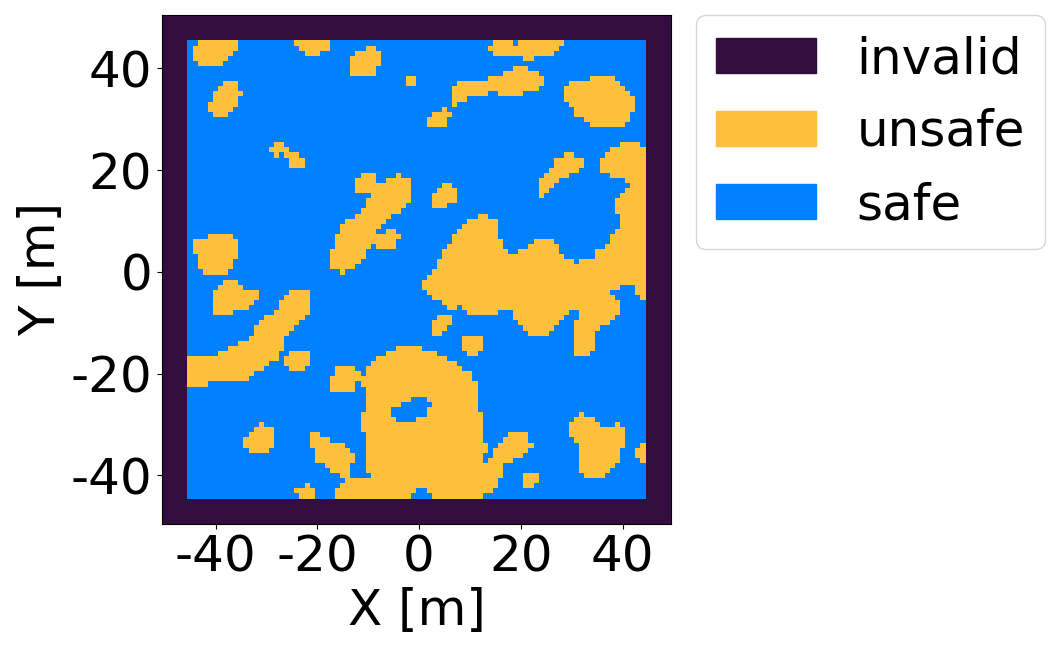}
    \caption{Prediction without uncertainty threshold} \label{fig:1c}
  \end{subfigure}
  \hspace*{\fill}   
  \begin{subfigure}{0.18\textwidth}
    \includegraphics[trim=0 0 0 0, clip, height=30mm]{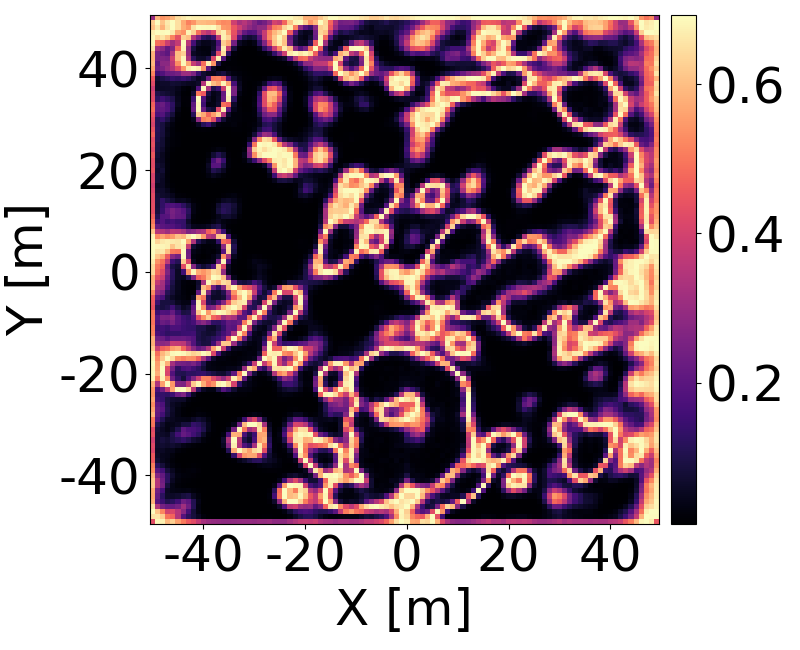}
    \caption{Uncertainty Map \newline} \label{fig:1d}
  \end{subfigure}

\caption{Example of safety map prediction and uncertainty estimation.}
\label{fig:prelim}
\end{figure}

\noindent Figure~\ref{fig:prelim} shows an example of the outputs.

From left to right, Figure \ref{fig:prelim} shows an example input noisy DEM, the ground truth label produced by running a replicated version of the ALHAT algorithm on an input true DEM without noise, the output network prediction, and the output uncertainty map. The network prediction shows the network output labeling each pixel as safe or unsafe, reported in blue and yellow, respectively. Note that border pixels are considered invalid as the safety assessment of each pixel requires consideration of neighboring pixels, with the size of the neighborhood dependent on the lander size. These pixels are ignored during training and evaluation. The uncertainty map is obtained by calculating predictive entropy. Low uncertainty, reported in black, indicates that the network prediction can be trusted. High uncertainty, reported in yellow, indicates that the network output may be inaccurate.

\subsection{Uncertainty-aware Safety Map Generation} \label{site_selection}

\begin{figure}[hbt!]
\centering
\includegraphics[width=.8\textwidth]{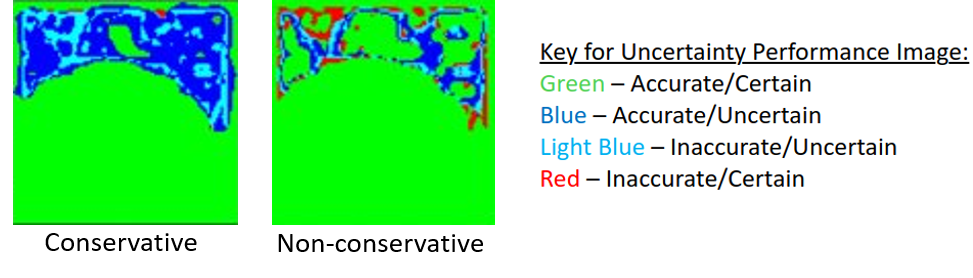}
\caption{Effect of varying the uncertainty threshold to determine if the prediction is certain or uncertain.}
\label{fig:threshold}
\end{figure}

It is necessary to select an uncertainty threshold, $\mathbbm{H}_t$, for determining whether the uncertainty value should be interpreted as certain or uncertain~\cite{Mukhoti2018}. Predictions determined to be uncertain should not be trusted. Formally, given an uncertainty threshold, $\mathbbm{H}_t$, and uncertainty value, $\hat{\mathbbm{H}}$, we interpret the prediction as uncertain and make it invalid if $\hat{\mathbbm{H}} \geq \mathbbm{H}_t$. Otherwise, the predicted labels are retained. Figure~\ref{fig:threshold} provides an illustrative example to demonstrate the effect of varying the uncertainty threshold to determine whether predictions for each pixel are certain or uncertain. Selecting a conservative threshold leads to many accurate pixels being labeled uncertain, potentially discarding accurate predictions. With a less conservative threshold, more predictions pass through as certain and accurate. However, some predictions appear as certain although they are inaccurate, which could be dangerous for a safety-critical application. Selecting the proper uncertainty threshold is a critical consideration for uncertainty-aware safety prediction.

\begin{figure}[hbt!]
\centering
\includegraphics[width=.90\textwidth]{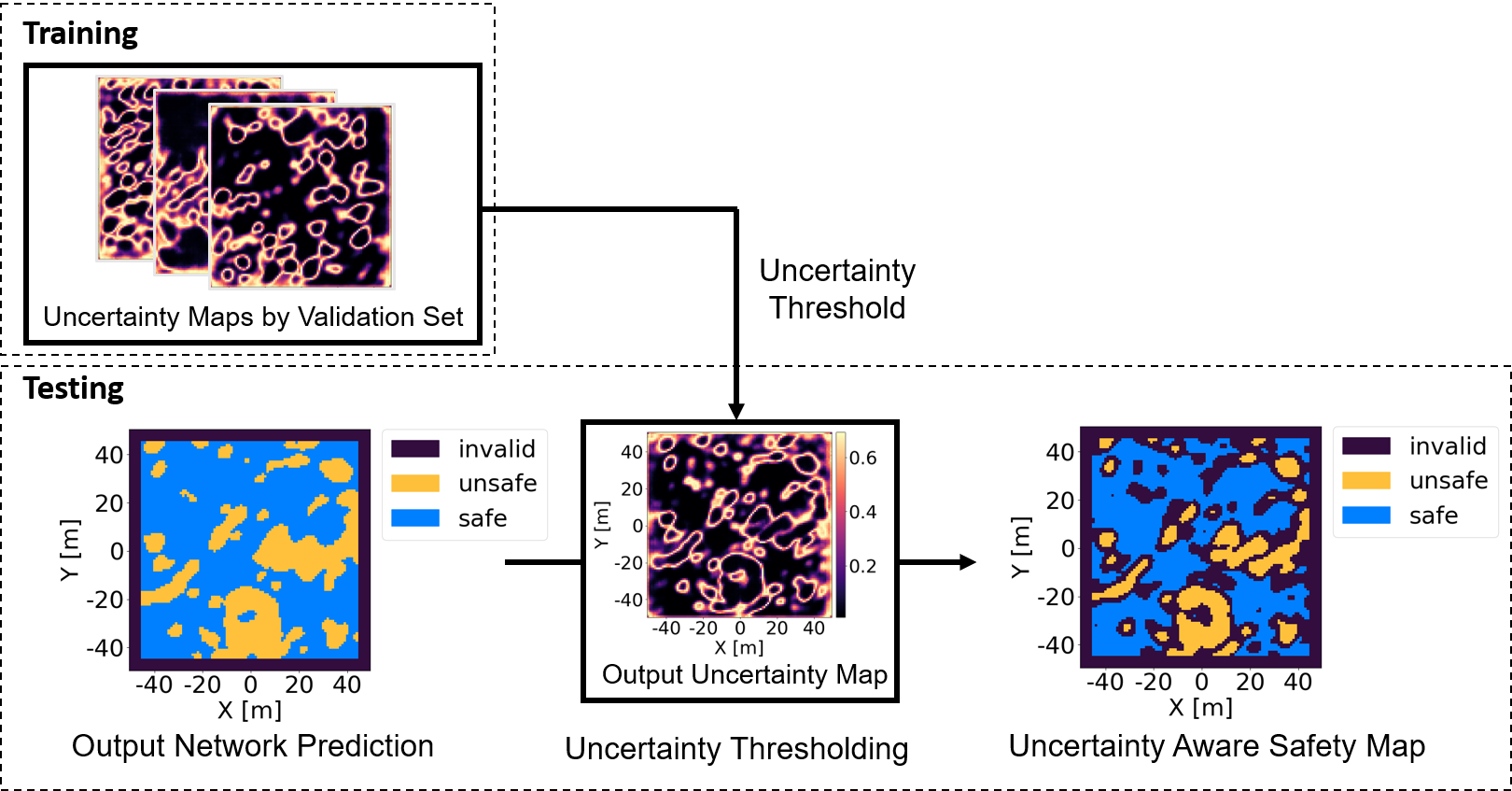}
\caption{Pipeline for integrating uncertainty map and network prediction to enable uncertainty-aware safety map generation.}
\label{fig:uncertaintypipeline}
\end{figure}

The uncertainty threshold is selected based on a statistic from the training set or, if the training set is too large, the validation set. Note that the uncertainty threshold should not be fully dependent on the information at the observation. For example, consider the case where the threshold is the mean value of the obtained uncertainty map. Then, within the predicted safety map, there always (incorrectly) exist some pixels interpreted certain, no matter how large the minimum value of the obtained uncertainty map is. In this study, the uncertainty threshold is set to be the mean uncertainty value across the validation set, as suggested by \citet{Mukhoti2018} for general semantic segmentation tasks. Moreover, we also studied the effect of varying the uncertainty threshold for our application in Section \ref{vary_ut}. Figure \ref{fig:uncertaintypipeline} summarizes the pipeline of integrating the uncertainty map and network prediction.


\section{Experiments \& Results}
\subsection{Dataset Preparation}
Network training and testing is achieved using simulated terrain data for input DEMs and a replicated version of the ALHAT's hazard detection (HD) algorithm for output safety maps.

\subsubsection{Simulated terrain data:}

\begin{figure}[htb]
  \centering
    \includegraphics[height=50mm]{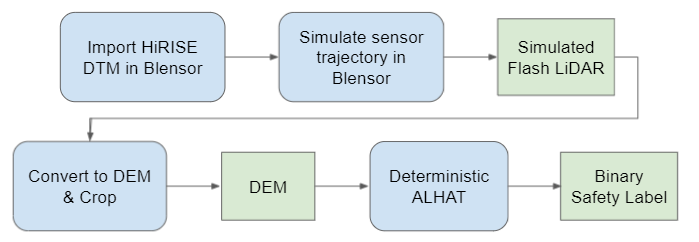}
    \caption{Flowchart illustrating the pipeline for creating a realistic virtual environment and generating simulated data.}
    \label{fig:simdata}
\end{figure}

\begin{figure}[htb]
 \begin{minipage}{0.40\hsize}
  \centering
    \includegraphics[height=60mm]{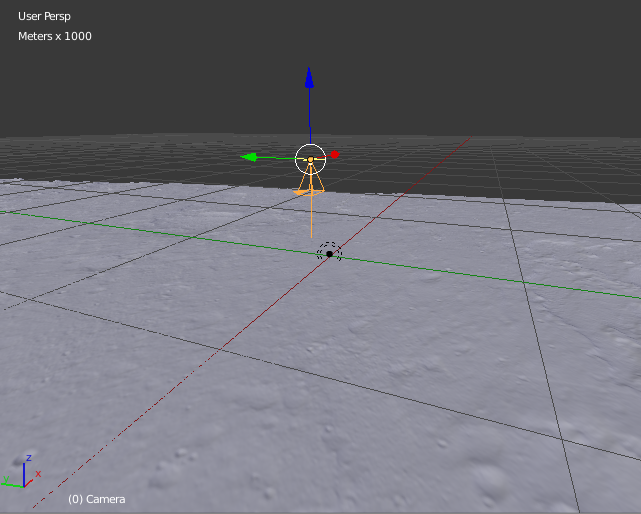}
    \caption{Simulation environment for data generation}
    \label{fig:blenderenvironment}
 \end{minipage}
 \begin{minipage}{0.60\hsize}
  \centering
   \includegraphics[trim=20 30 10 30, clip, height=80mm]{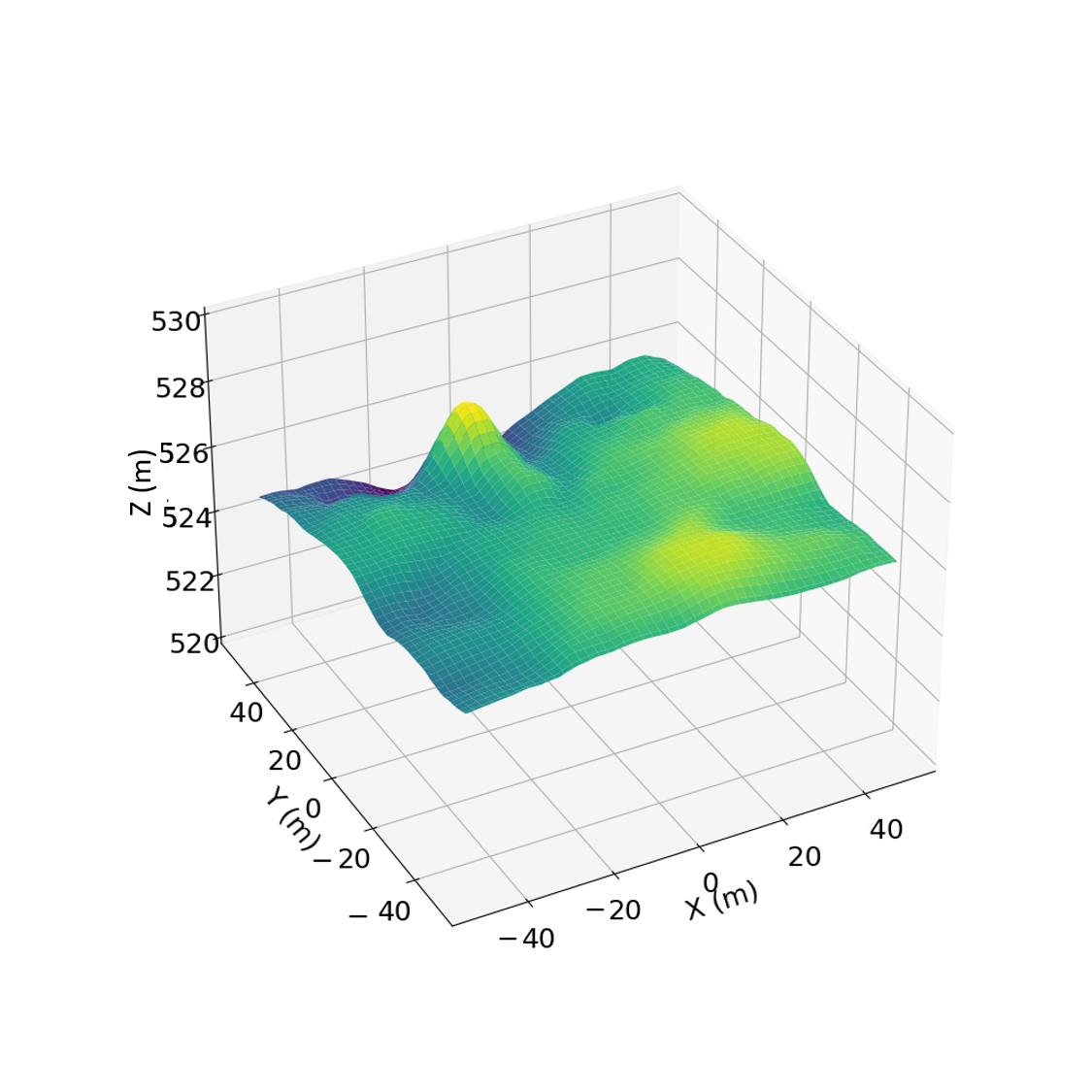}
  \caption{Example of generated terrain}
  \label{fig:terrain_sample}
 \end{minipage}
\end{figure}

Improving from our previous work~\citep{Skinner2021}, this work leverages an open-source sensor simulation toolbox, BlenSor, to simulate data collection of realistic terrain with LiDAR~\citep{Gschwandtner2011}. BlenSor is built upon Blender, an open-source 3D simulation engine. Figure~\ref{fig:simdata} illustrates the pipeline for creating the simulation environment and generating the simulated dataset. To create the simulation environment, we first load a Mars High Resolution Imaging Science Experiment (HiRISE) digital terrain model (DTM) into BlenSor using the Blender HiRISE plug-in~\cite{McEwen2007}. The Mars HiRISE DTMs are derived from stereo image pairs collected by the HiRISE camera on the Mars Reconnaissance Orbiter (MRO)~\citep{McEwen2007}. This provides a realistic base terrain collected during a real mission for generating simulated data. The original Mars HiRISE DTM is selected for the candidate future landing site near Sabrina Vallis with $1$ m/pixel resolution. Due to memory limitations, we load the DTM at $25$\% of the full resolution. Next, we configure a time-of-flight sensor to simulate the parameters of a real landing scenario. Table~\ref{tab:sensor} provides the sensor parameters of the simulated sensor. These parameters are selected to obtain DEMs with similar dimensions and characteristics of DEMs obtained through real and simulated experiments with flash LiDAR surveys for imaging planetary surfaces~\citep{Restrepo2020, Keim2010, Bulyshev2011, Amzajerdian2016}. For the simulation, the sensor is moved across the terrain at a fixed altitude of $500$ m to collect incremental scans of the terrain. The output from BlenSor is a point cloud. The point cloud is converted to a DEM with a resolution of $1$ m/pixel using bilinear interpolation. The final DEM is cropped to a resolution of $100$ x $100$. BlenSor allows for the output of the true (noise-free) sensor scan, or a noisy scan. We generated a low-noise dataset with a 1-$\sigma$ noise level of 1.67 cm (S-167), a moderate-noise dataset with a 1-$\sigma$ noise level of 3 cm (S-300), and a high-noise dataset with a 1-$\sigma$ noise level of 7 cm (S-700). Each dataset uses the same terrain and the same sensor trajectory to ensure that the main difference in the datasets is the noise level. Figure~\ref{fig:blenderenvironment} shows the simulation environment during the data generation process, and Fig. \ref{fig:terrain_sample} shows the sample output terrain obtained through our simulation pipeline. 

Note that we can add any other noise models like distortion or damaged elements of the LiDAR detector for the training data, given the mission-specific observation model. The altitudes and the angles of the sensor boresight axis can also be varied so that the training data distribution aligns well with a specific situation during descent. In this paper, however, we assume that we can perform the appropriate resampling of the raw DEM from the LiDAR point cloud data (PCD). This process up-samples or down-samples the raw DEM from the PCD to have a uniform resolution that is identical to the training DEMs. The detailed resampling-based approach and its application in a simulated Mars landing scenario can be found in Ref. \cite{tomita2022adaptive}.

\begin{table}[htbp!]
\caption{Sensor parameters for generating simulated data in BlenSor~\cite{Gschwandtner2011}.}
\begin{center}
\begin{tabular}{ c c }
\hline
 \textbf{Parameter} & \textbf{Value}  \\ \hline \hline
Detector size & 128x128 \\  \hline
Horizontal field-of-view & $12^{\circ}$ \\ \hline
Vertical field-of-view & $12^{\circ}$ \\ \hline
Focal length & 25 cm \\ \hline
\end{tabular}
\label{tab:sensor}
\end{center}
\end{table}

\subsubsection{Label generation:}
To prepare ground truth labels of safety maps, we replicated ALHAT's HD algorithm \citep{Ivanov2013}. ALHAT's HD algorithm calculates the maximum slope and roughness the lander would experience at touchdown by explicitly evaluating the lander geometry and the landing pad contacts to the given terrain. We replicated deterministic slope evaluation, deterministic roughness evaluation, and probabilistic roughness evaluation. For label generation, we applied deterministic slope and roughness evaluation to the noise-free DEMs. For ALHAT performance evaluation as a baseline, we applied deterministic slope evaluation and probabilistic roughness evaluation as proposed by \citep{Ivanov2013}, resulting in the algorithm that returns pixel-wise safety probability. The baseline performance is measured by adopting a safety threshold of 50\% to assign a label of safe or unsafe to each pixel. Note that this algorithm requires evaluation of all the surrounding pixels for every aiming point, so it may not be used online for high-resolution DEM inputs without approximation. 

\subsection{Implementation Details}
We generated simulated datasets of 1000 DEMs. From this data, 800 DEMs are used for training, 100 DEMs are used for validation, and 100 DEMs are used for testing, selected through random selection. Each training set contains a noised DEM and a label generated with the replicated ALHAT algorithm from the noise-free DEM. The input data is 100x100 resolution at 1 meter per pixel. The data is upsampled with bilinear interpolation to 512x512 before being input to the network. The data is also normalized between 0 and 1 based on the minimum and maximum values across the training set. 

For training, we use a batch size of 8. The dropout rate for dropout layers is set to 0.5 following Reference~\citenum{Kendall2015}. The learning rate is 0.0001 and the momentum is 0.9. The network is trained for $10^4$ epochs, which means that the full training set is seen $10^4$ times. This training takes approximately 2 days on an NVIDIA GeForce RTX 2070 SUPER GPU. For testing, we use $M=8$ samples for MC-dropout.

\subsection{Results \& Discussion}
\subsubsection{Qualitative Comparison with Visualized Output}

\begin{figure}[htbp!]
\centering

\begin{tabular}{c c}
    \includegraphics[trim=0 0 0 0, clip, height=40mm]{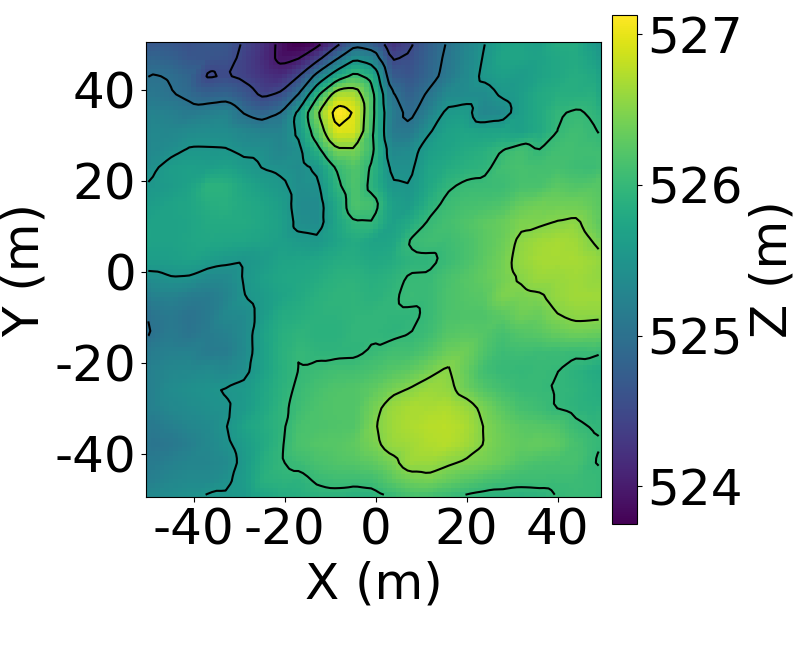} &
    \includegraphics[trim=0 0 0 0, clip, height=40mm]{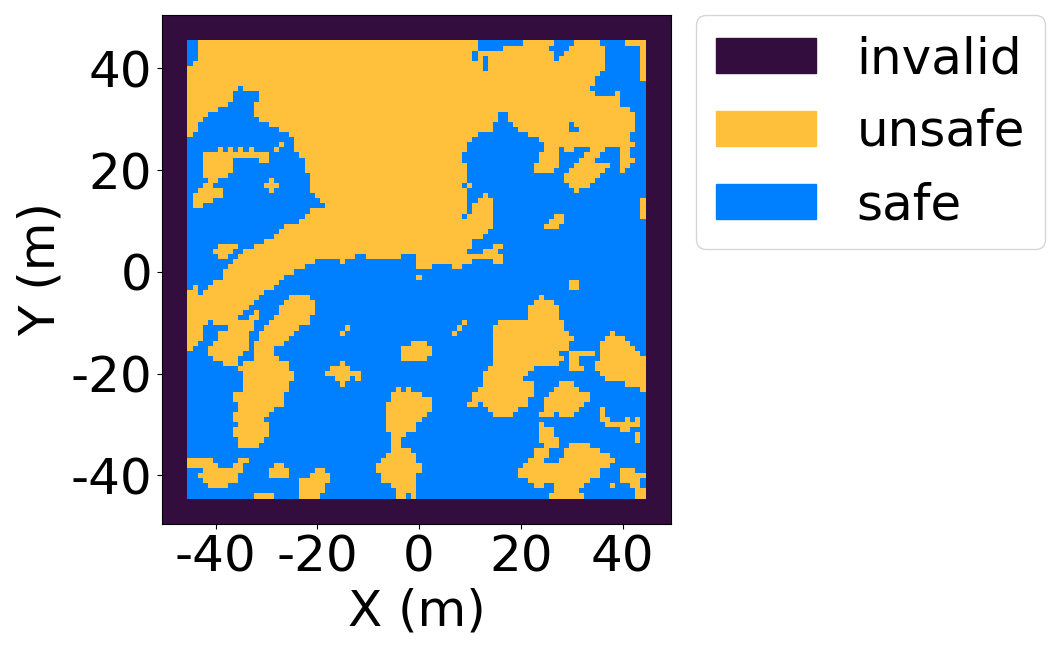} \\
    \centering (a) True DEM & \centering (b) Ground Truth 
    \vspace{3mm}
\end{tabular}

\begin{tabular}{p{10mm} p{32mm} p{35mm} p{30mm} p{30mm}}
    S-167: 1$\sigma=$ 1.67cm & 
    \begin{minipage}{.4\textwidth}
    \includegraphics[trim=0 0 137 0, clip, height=35mm]{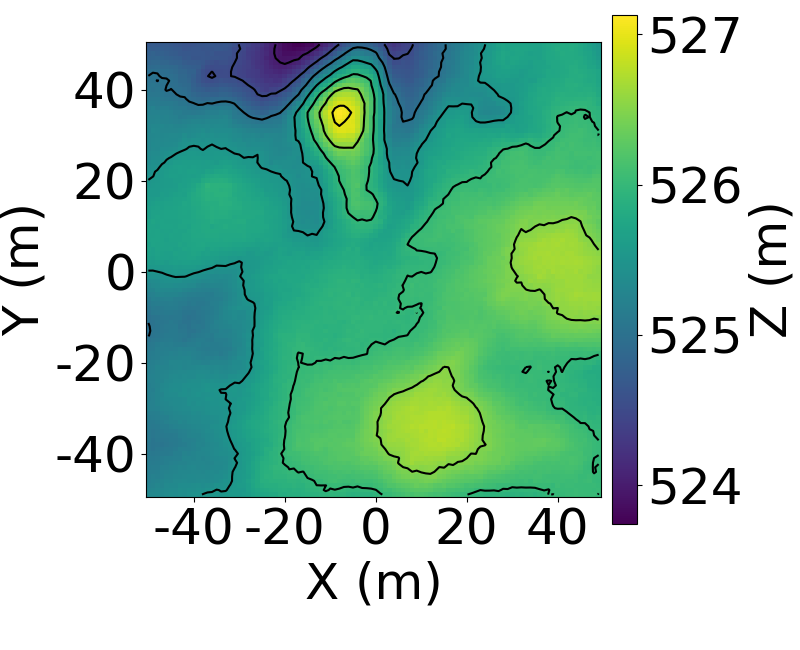}
    \end{minipage} &
    \begin{minipage}{.4\textwidth}
    \includegraphics[trim=50 0 0 0, clip, height=32mm]{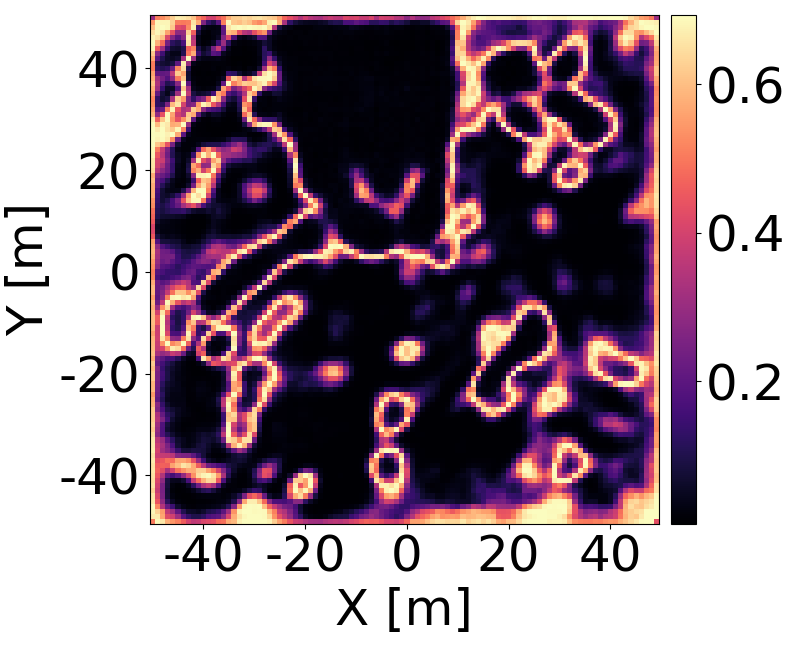} 
    \end{minipage}&
    \begin{minipage}{.4\textwidth}
    \includegraphics[trim=50 0 260 0, clip, height=32mm]{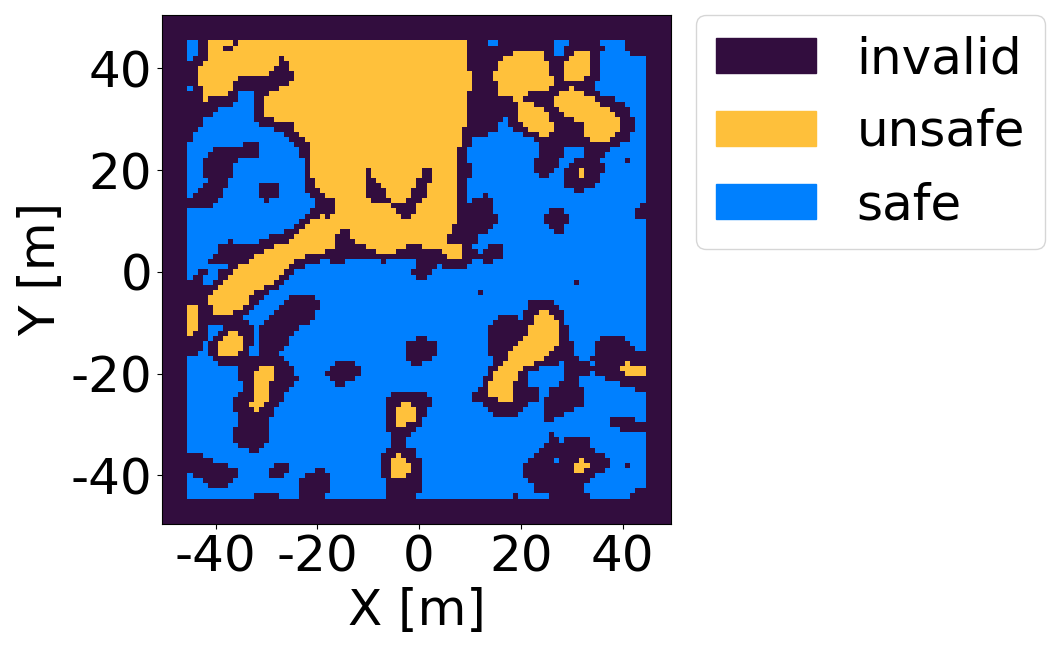} 
    \end{minipage}&
    \begin{minipage}{.4\textwidth}
    \includegraphics[trim=50 0 260 0, clip, height=32mm]{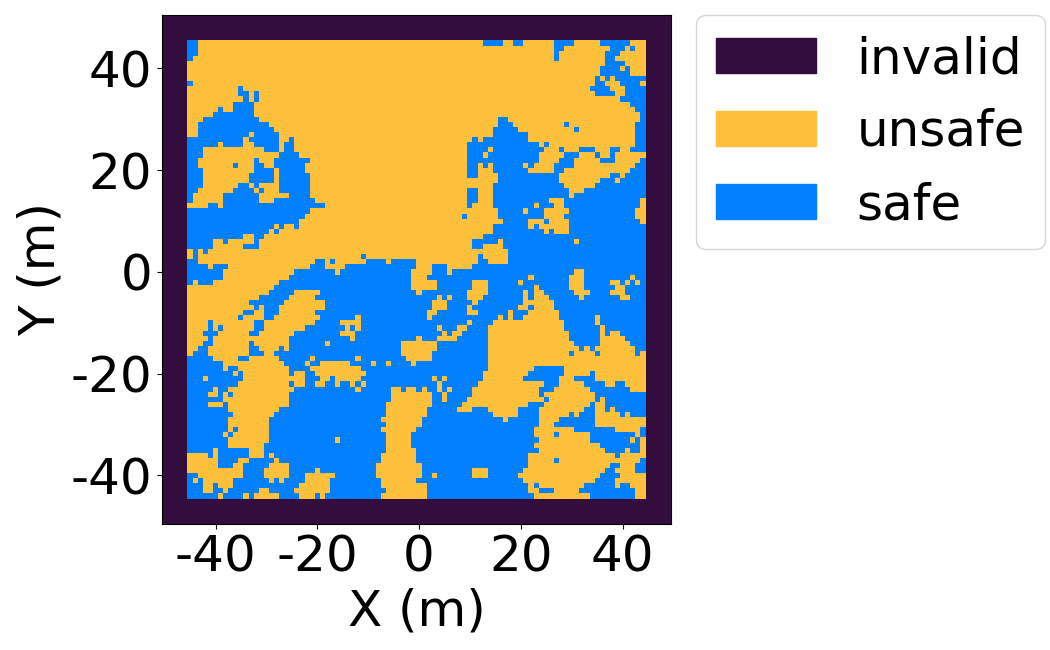} 
    \end{minipage}\\
    
    S-300: 1$\sigma=$ 3.00cm & 
    \begin{minipage}{.3\textwidth}
    \includegraphics[trim=0 0 137 0, clip, height=35mm]{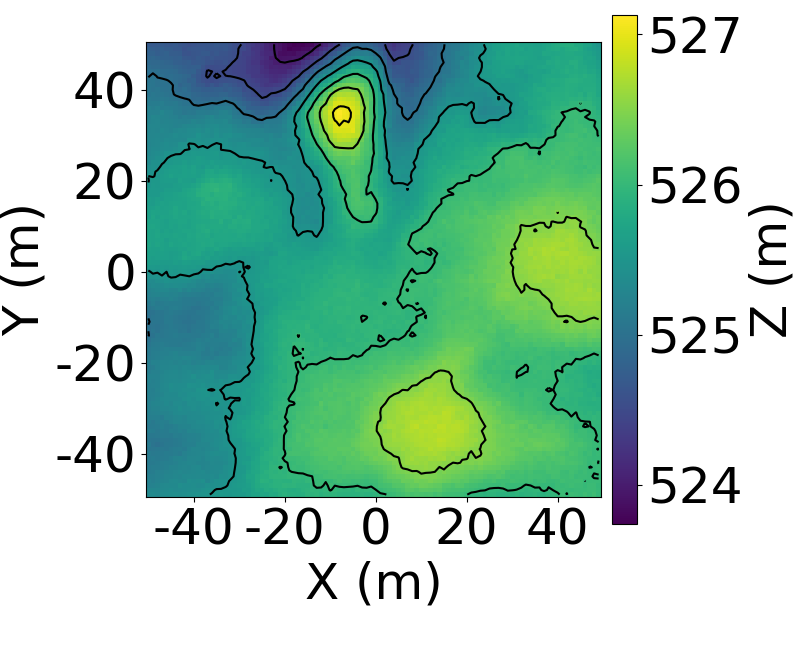} 
    \end{minipage}&
    \begin{minipage}{.3\textwidth}
    \includegraphics[trim=50 0 0 0, clip, height=32mm]{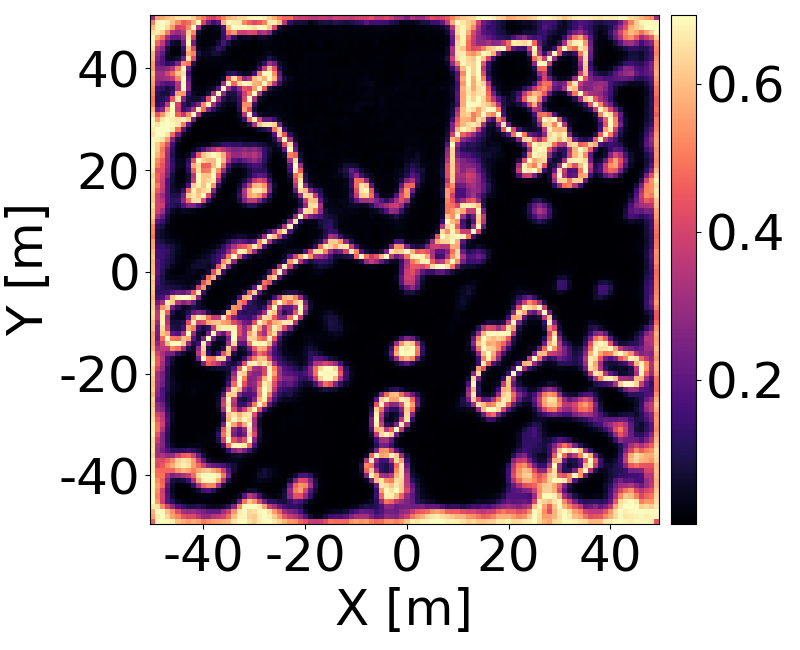} 
    \end{minipage}&
    \begin{minipage}{.3\textwidth}
    \includegraphics[trim=50 0 260 0, clip, height=32mm]{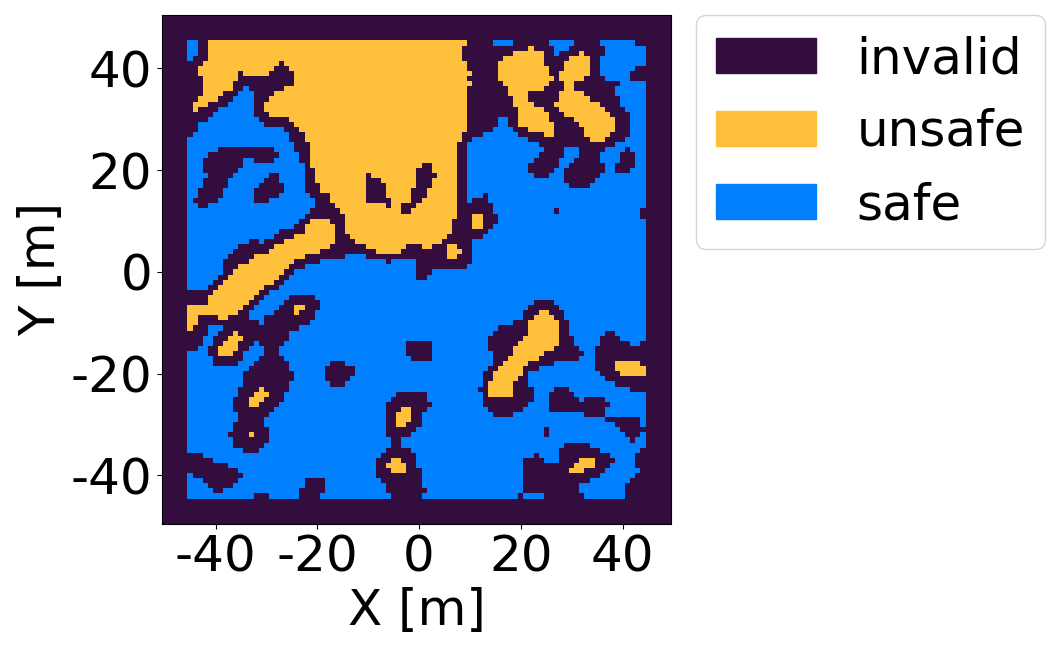} 
    \end{minipage}&
    \begin{minipage}{.3\textwidth}
    \includegraphics[trim=50 0 260 0, clip, height=32mm]{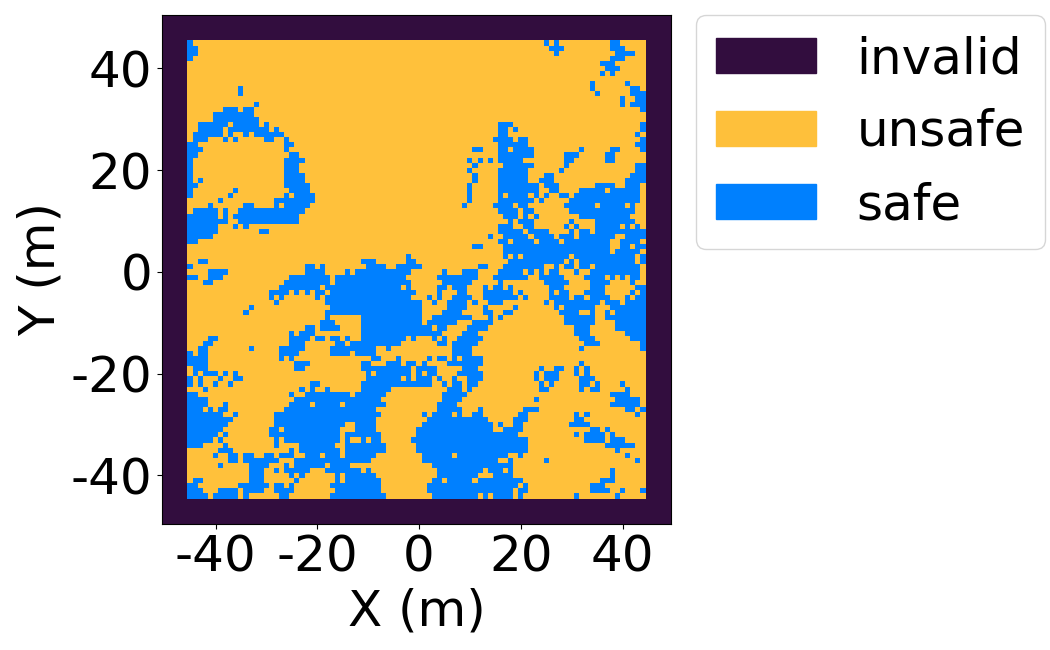} 
    \end{minipage}\\
    
    S-700: 1$\sigma=$ 7.00cm & 
    \begin{minipage}{.3\textwidth}
    \includegraphics[trim=0 0 137 0, clip, height=35mm]{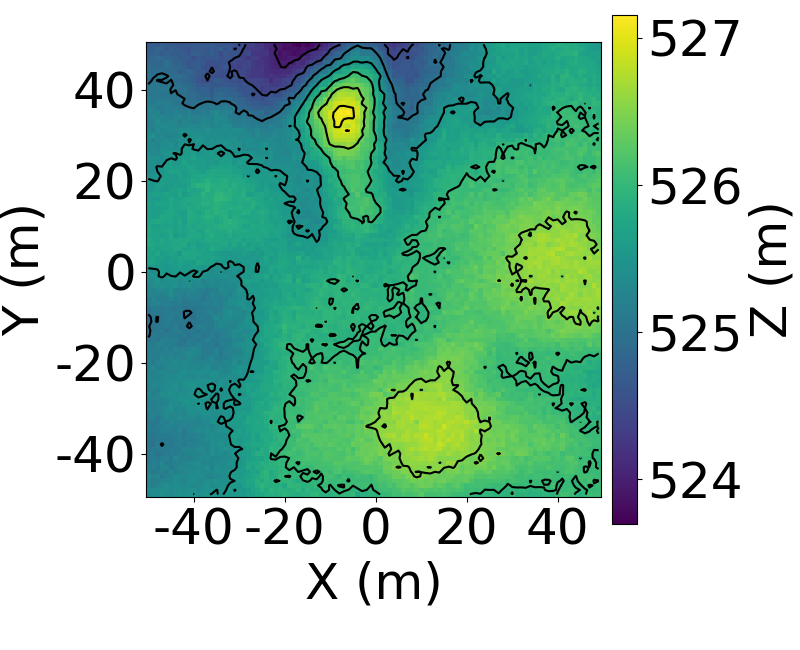} 
    \end{minipage}&
    \begin{minipage}{.3\textwidth}
    \includegraphics[trim=50 0 0 0, clip, height=32mm]{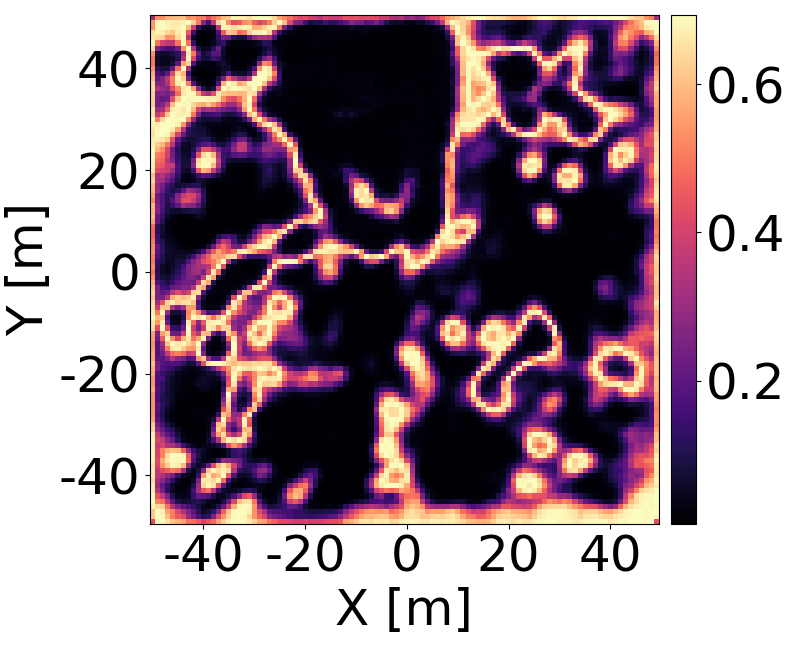} 
    \end{minipage}&
    \begin{minipage}{.3\textwidth}
    \includegraphics[trim=50 0 260 0, clip, height=32mm]{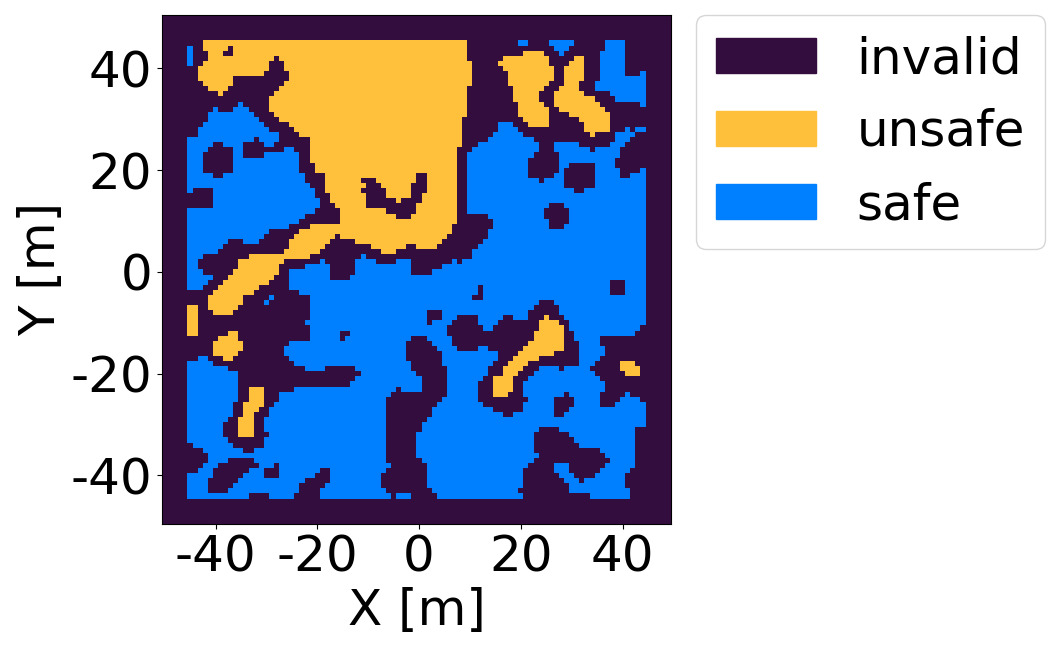} 
    \end{minipage}&
    \begin{minipage}{.3\textwidth}
    \includegraphics[trim=50 0 260 0, clip, height=32mm]{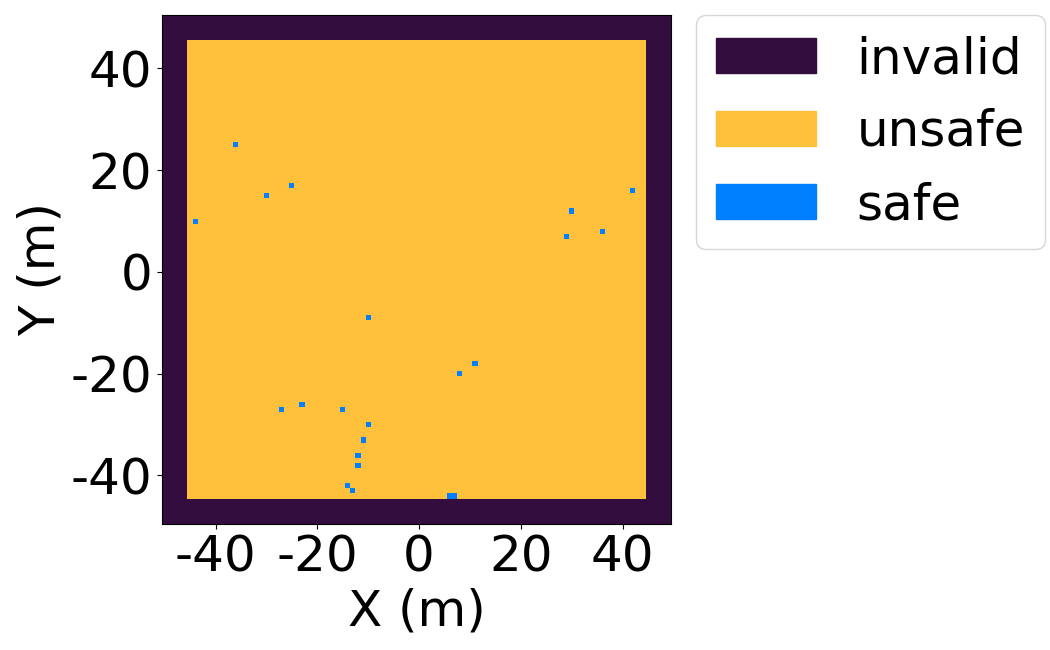} 
    \end{minipage}\\
        
    & \centering (c) Noised DEM & \centering (d) Uncertainty Map & \centering (e) Uncertainty Aware Prediction & \centering (f) Baseline \\
      
\end{tabular}
\caption{Qualitative sample results compared with baseline method for varying levels of input noise.}
\label{fig:qualresults}
\end{figure}

Figure~\ref{fig:qualresults} shows qualitative results of the network prediction together with the baseline performance for a sample DEM with varying levels of noise: low noise of 1.67 cm in 1-$\sigma$ (S-167), moderate-noise of 3 cm in 1-$\sigma$ (S-300), and high-noise of 7 cm in 1-$\sigma$ (S-700). Note that the uncertainty threshold is selected to be the mean uncertainty in the S-167 validation set for all cases. On the top, we show the true, noise-free DEM and corresponding safety label. We can observe that safe-unsafe border pixels have higher uncertainty, and their area increases as the noise level increases. As our method incorporates the uncertainty map in our final prediction, our uncertainty-aware safety map has decreased error for valid pixels, at the expense of increasing the number of invalid pixels. In other words, our method can cope with the increased sensor noise with the aid of the uncertainty map. On the other hand, the baseline prediction from the replicated ALHAT algorithm, which is a state-of-the-art method, degrades with increased sensor noise as reported in Ivanov et al.~\citep{Ivanov2013}. The baseline method is conservative, especially for the high noise input, where all pixels are labeled as unsafe and there are no viable safe landing sites. The learning-based approach can still return viable safe landing site candidates, even when the noise of the test sample is outside the distribution of the training set.

\subsubsection{Quantitative Comparison with Performance Metrics}
The following section presents quantitative metrics to evaluate the results: pixel accuracy, mean intersection over union, precision, and sensitivity. Each metric is described below.

Pixel accuracy, PA, is computed by

\begin{align}
\textrm{PA} = \frac{\sum_{i=1}^{n_c} N_{ii}}{\sum_{i=1}^{n_c} \sum_{j=1}^{n_c} N_{ij}}
\end{align}

\noindent where $n_c = |\mathcal{C}|$ is the number of classes, and $N_{ij}$ is the number of pixels of class $i$ predicted as class $j$~\cite{Long2015}. We compute pixel accuracy across all of the valid pixels in the test set. Invalid pixels are the border pixels and, for the proposed uncertainty-aware method, the pixels with higher uncertainty than the threshold.

Mean intersection over union is a common metric for evaluating semantic segmentation. Intersection over union for a class $c$ is computed by

\begin{align}
    \textrm{IoU}(c) = \frac{N_{cc}}{\left(\sum_{i=1}^{n_c} N_{ci} + \sum_{i=1}^{n_c} N_{ic}\right) - N_{cc}}
\end{align}

\noindent where the numerator and denominator represent respectively the intersection and union of the predicted pixels and the ground truth of the class $c$ pixels. The mean intersection over union (mIoU) is defined as the mean IoU across all classes, $\textrm{mIoU}={n_c}^{-1}\sum_{c \in \mathcal{C}}\textrm{IoU}(c)$. We compute mIoU with valid pixels only, as in the PA evaluation.

Lastly, we provide precision and sensitivity. Precision represents how reliable the predicted safe sites are and sensitivity represents how many safe sites are detected among the true safe sites. These values are calculated as following:

\begin{align}
    \textrm{Precision} = \frac{\textrm{(True Safe)}}{\textrm{(True Safe) + (False Safe)}}
\end{align}
\begin{align}
    \textrm{Sensitivity} = \frac{\textrm{(True Safe)}}{\textrm{(True Safe) + (False Unsafe)}}
\end{align}

\noindent where false unsafe include the safe pixels in ground truth that are classified as invalid due to their high uncertainty. These counts are accumulated across pixels for the full test set and the rates are computed at the end.

Table~\ref{table:results} shows the results for these metrics compared across different methods. The baseline method results are generated from our replicated version of the probabilistic ALHAT algorithm. The base network method results, denoted by "Base Net.", evaluate the initial network prediction obtained through MC-sampling with Eq. 1, which is the output safety map without incorporating uncertainty. The uncertainty-aware prediction is the result of our proposed method to integrate estimated uncertainty and the network prediction to provide a final uncertainty-aware safety map. The uncertainty threshold is selected to be the mean uncertainty in the validation set of the S-167 dataset.

In Table~\ref{table:results}, the baseline achieves the best precision value across the various noise level inputs, but its sensitivity significantly degrades as the noise increases. In other words, the baseline is too conservative to detect safe sites by the severely noised DEMs. This result is consistent with the visualization in Fig.~\ref{fig:qualresults}. In contrast, the base network detects the safe sites most as seen in the best sensitivity values, while the precision is lowered. The proposed method achieves the balanced precision and sensitivity values and also the best overall segmentation performance with the best pixel accuracies and mIoUs. The rate of pixels that pass the uncertainty thresholding, shown in the V/C Pix. \% column, decreases as the noise level increases. However, the other performance metrics of the base network and the proposed uncertainty-aware methods do not degrade as much as the baseline method.

\begin{table}[htbp!]
\centering
\caption{Quantitative results comparison between the baseline replicated ALHAT method (Baseline), Bayesian deep learning prediction before uncertainty thresholding (Base Net.), and uncertainty-aware Bayesian deep learning prediction (Uncertainty-Aware). V/C pixels stand for valid and certain pixels and Pix. Acc. stands for pixel accuracy. Bold numbers correspond to the best results for each combination of the noise case and the metric.}
\begin{tabular}{p{30mm}p{10mm}p{10mm}p{9mm}cccc}
\hline
 \textbf{Method} & \textbf{Train Noise 1-$\sigma$} &  \textbf{Test Noise 1-$\sigma$} & \textbf{V/C Pix.\%} &  \textbf{Pix. Acc.} &  \textbf{mIoU} & \textbf{Precision} & \textbf{Sensitivity} \\ \hline \hline
 
 Baseline \newline (Replicated ALHAT) &  -- & 1.67cm & -- & 0.9345 & 0.8294 & \textbf{0.9798} & 0.7506 \\ \hline
 Base Net.  &  1.67cm & 1.67cm & -- & 0.8687  & 0.7347 & 0.6639 & \textbf{0.9488} \\ \hline
 {Uncertainty-Aware} &  1.67cm & 1.67cm & 64\% & \textbf{0.9839}$^*$ & \textbf{0.9615}$^*$ & 0.9493 & 0.7182 \\ \hline \hline
 
 Baseline \newline (Replicated ALHAT)&  -- & 3cm & -- & 0.8702 & 0.6663 & \textbf{0.9795} & 0.4849 \\ \hline
 Base Net. &  1.67cm & 3cm & -- & 0.8516 & 0.7095 & 0.6330 & \textbf{0.9510} \\ \hline
{Uncertainty-Aware} &  1.67cm & 3cm & 63\% & \textbf{0.9712}$^*$ & \textbf{0.9343}$^*$ & 0.9228 & 0.7375 \\ \hline \hline
 
  Baseline \newline (Replicated ALHAT)&  -- & 7cm & -- & 0.7539 & 0.3790 & \textbf{0.9723} & 0.0044\\ \hline
 Base Net. &  1.67cm & 7cm & -- & 0.8487 & 0.7032 & 0.6321 & \textbf{0.9284} \\ \hline
  {Uncertainty-Aware} &  1.67cm & 7cm & 58\%  & \textbf{0.9722}$^*$ & \textbf{0.9320}$^*$ & 0.9092 & 0.5934 \\ \hline

\end{tabular}
\label{table:results}
\begin{tablenotes}
\item {\footnotesize $^*$ The metric is evaluated by ignoring any pixel that has higher uncertainty than the threshold. The V/C Pix.\% shows the rate that passed the uncertainty thresholding. The uncertainty threshold is selected to be the mean uncertainty in the S-167 validation set.}
\end{tablenotes}
\end{table}

\subsubsection{Effect of Varying the Uncertainty Threshold}\label{vary_ut}
\begin{figure}[htbp!]
\centering
\begin{tabular}{c c c}
    \includegraphics[trim=10 0 9 0, clip, height=40mm]{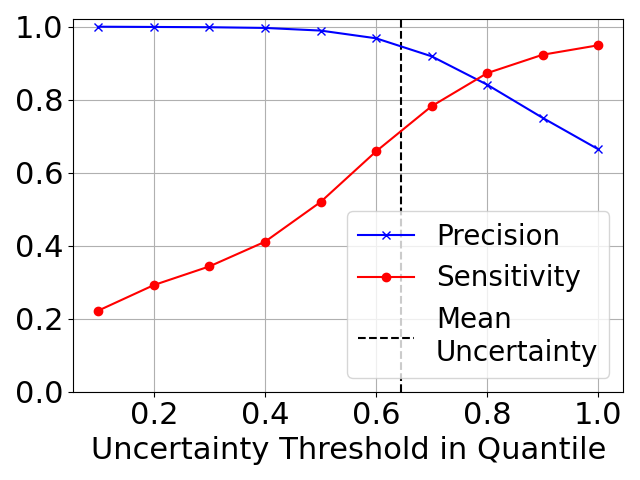} &
    \includegraphics[trim=10 0 9 0, clip, height=40mm]{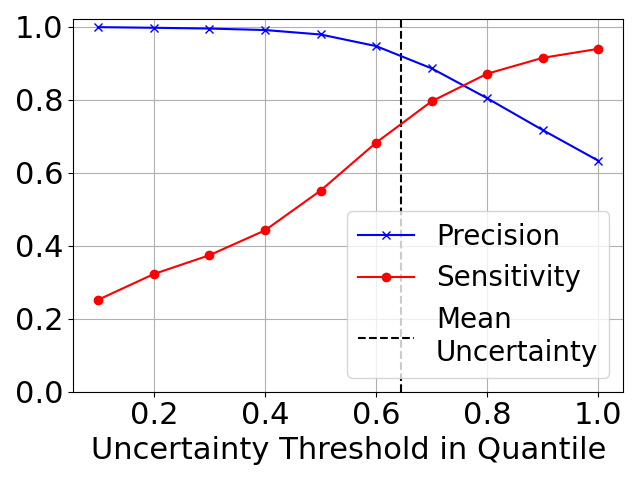} &
    \includegraphics[trim=10 0 9 0, clip, height=40mm]{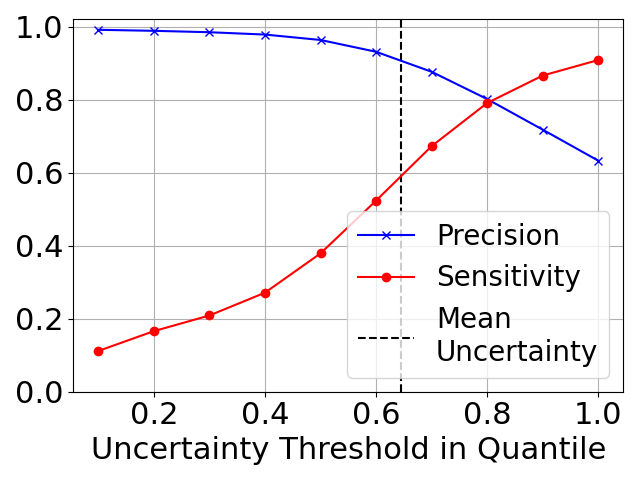} \\
    \centering (a) S-167: 1$\sigma=$ 1.67cm & \centering (b) S-300: 1$\sigma=$ 3.00cm & \centering (c) S-700: 1$\sigma=$ 7.00cm 
\end{tabular}
\caption{Qualitative sample results compared with baseline method for varying levels of input noise.}
\label{fig:uts}
\end{figure}

Figure~\ref{fig:uts} shows the precision and sensitivity of the proposed method with varying uncertainty thresholds for all three noise level cases. The selected uncertainty threshold is expressed in quantile in the S-167 validation set. For instance, the fifth data points from the left with the horizontal axis value of 0.5 represent when the uncertainty threshold is selected to be the median of the uncertainty values of the S-167 validation set. The dashed line denotes the mean uncertainty of the validation set, which is the threshold used in Fig.~\ref{fig:qualresults} and Table~\ref{table:results}. We can observe that the lower uncertainty threshold makes more pixels classified invalid, which results in lower sensitivity and less safe sites detected. In contrast, the lower threshold makes precision higher and the predicted safe sites more reliable. 

As a hazard detection algorithm for autonomous safe planetary landings, the goal is to provide as many safe sites as possible while maintaining the reliability of the prediction. Figure~\ref{fig:uts} shows we can achieve this by adjusting the uncertainty threshold accordingly. For instance, in the S-300 case, setting uncertainty threshold as the 40\% quantile achieves 99.05\% precision and 44.18\% sensitivity and 50\% quantile threshold with 97.83\% precision and 55.07\% sensitivity. Compared with the baseline of 97.95\% precision and 48.49\% sensitivity, the proposed method offers the flexibility to explore the trade-off between precision and sensitivity. In the S-700 case, the proposed method can outperform the baseline in terms of both precision and sensitivity. Setting the uncertainty threshold as the 40\% quantile in the S-700 case achieves the precision of 97.80\% while having the sensitivity of 27.12\%, which significantly outperforms the baseline with 97.23\% precision and 0.44\% sensitivity.

\subsubsection{Effect of Dead Elements of LiDAR Detector}
LiDAR systems might have a few completely dead elements, or a few damaged elements with large range errors. Such an incomplete observation also needs to be processed by the hazard detection algorithm without severely degrading the performance. Ideally, we would retrain the network with the damaged DEMs. However, our analysis shows that bilinear interpolation is sufficient for the proposed method to handle the incomplete observation when the missed range measurements are sparse and limited in number. Table~\ref{table:results_hole} shows the performance of each method for the simulated damaged DEMs. We randomly dropped ten sparse pixels out of the 100 x 100 input DEM pixels. The dropped pixels are reconstructed via bilinear interpolation before the upsampling to the 512 x 512 pixels and processed by the network. The metrics evaluation for all cases including baseline, ignores any pixel around the dropped pixel with the lander size, for which the information is insufficient and we avoid landing. The pixel accuracy and mIoU for the proposed method in the third row of Table~\ref{table:results_hole} ignores also the pixels with higher uncertainty than the threshold. The uncertainty threshold is selected to be the mean uncertainty value in the validation set of S-167, which does not contain the damaged DEMs. The results in Table~\ref{table:results_hole} show that the dropped pixels in the damaged DEMs do not affect the prediction by the proposed method when the DEM is corrected via bilinear interpolation and the minimum neighborhood of those pixels by the lander size is ignored.

\begin{table}[htbp!]
\centering
\caption{Quantitative results comparison for the restored input DEMs from the incomplete observation with ten randomly dropped pixels, between the baseline replicated ALHAT method (Baseline), Bayesian deep learning prediction before uncertainty thresholding (Base Net.), and uncertainty-aware Bayesian deep learning prediction (Uncertainty-Aware). V/C pixels stand for valid and certain pixels and Pix. Acc. stands for pixel accuracy. Bold numbers correspond to the best results for each combination of the noise case and the metric.}
\begin{tabular}{p{30mm}p{10mm}p{10mm}p{9mm}cccc}
\hline
 \textbf{Method} & \textbf{Train Noise 1-$\sigma$} &  \textbf{Test Noise 1-$\sigma$} & \textbf{V/C Pix.\%} &  \textbf{Pix. Acc.} &  \textbf{mIoU} & \textbf{Precision} & \textbf{Sensitivity} \\ \hline \hline
 
 Baseline \newline (Replicated ALHAT) &  -- & 1.67cm & -- & 0.9350$^*$ & 0.8299$^*$ & \textbf{0.9796}$^*$ & 0.7511$^*$ \\ \hline
 Base Net.  &  1.67cm & 1.67cm & -- & 0.8687$^*$  & 0.7308$^*$ & 0.6715$^*$ & \textbf{0.9122}$^*$ \\ \hline
 {Uncertainty-Aware} &  1.67cm & 1.67cm & 63\% & \textbf{0.9828}$^{{\dagger}}$ & \textbf{0.9567}$^{{\dagger}}$ & 0.9488$^*$ & 0.6517$^*$ \\ \hline \hline
 
\end{tabular}
\label{table:results_hole}
\begin{tablenotes}
\item {\footnotesize $^*$ The metric is evaluated by ignoring any pixel neighboring the dropped pixel with the size of the neighborhood dependent on the lander size.}
\item {\footnotesize $^{\dagger}$ The metric is evaluated by ignoring any pixel that has higher uncertainty than the threshold or is neighboring the dropped pixel with the size of the neighborhood dependent on the lander size. The V/C Pix.\% shows the rate that passed the uncertainty thresholding. The uncertainty threshold is selected to be the mean uncertainty in the S-167 validation set.}

\end{tablenotes}
\end{table}


\section{Conclusion}
This paper proposed a method for uncertainty-aware deep learning for safe landing site detection on planetary surfaces. Despite the recent advances of artificial intelligence techniques, their applicability to safety-critical space missions has often been limited due to concerns regarding their outputs' reliability. In this paper, we demonstrated that incorporating network uncertainty into the final safety map can improve its performance of safe landing site detection. More importantly, we also demonstrated that safe sites can be reliably detected from DEMs with high noise, and from DEMs where the sensor noise of the test samples is outside the distribution of the training samples. The proposed method can be an important step towards a reliable and robust machine-learning method for realistic spacecraft applications. 


\section*{Acknowledgments}
This work is supported by the National Aeronautics and Space Administration under Grant No.80NSSC20K0064 through the NASA Early Career Faculty Program.

\bibliography{references}

\end{document}